\documentclass{article}
\usepackage{spconf,amsmath,graphicx}
\usepackage{url}
\usepackage{amsmath}
\usepackage[linesnumbered,ruled]{algorithm2e}

\newlength\mylen
\newcommand\myinput[1]{%
  \settowidth\mylen{\KwIn{}}%
  \setlength\hangindent{\mylen}%
  \hspace*{\mylen}#1\\}
\let\oldnl\nl% Store \nl in \oldnl
\newcommand{\nonl}{\renewcommand{\nl}{\let\nl\oldnl}}% Remove line number for one line

\usepackage{hhline}
\usepackage{multirow}
\usepackage{arydshln}
\usepackage{float}

\usepackage[dvipsnames]{xcolor}
\definecolor{_pink}{rgb}{0.858, 0.188, 0.478}

\title{Random Shadows and Highlights: \\ A new data augmentation method for Extreme Lighting Conditions}
\name{Osama Mazhar, Jens Kober\thanks{This research is funded by H2020 EU project OpenDR, Grant Agreement No. 871449.}}
\address{Cognitive Robotics Department\\
        Delft University of Technology \\
        Delft, The Netherlands}
\begin{document}
%\ninept
%
\maketitle
\begin{abstract}
% Lately, deep neural networks are claimed to have achieved unprecedented performance on tasks including image classification, face identification and object detection.
% However, the generalization property of these networks across changes in the input distribution such as illumination changes and harsh conditions is uncertain.
In this paper, we propose a new data augmentation method, Random Shadows and Highlights (RSH) to acquire robustness against lighting perturbations.
Our method creates random shadows and highlights on images, thus challenging the neural network during the learning process such that it acquires immunity against such input corruptions in real world applications.
It is a parameter-learning free method which can be integrated into most vision related learning applications effortlessly.
With extensive experimentation, we demonstrate that RSH not only increases the robustness of the models against lighting perturbations, but also reduces over-fitting significantly.
Thus RSH should be considered essential for all vision related learning systems.
Code is available at: \textcolor{_pink}{ \url{https://github.com/OsamaMazhar/Random-Shadows-Highlights}}.
\end{abstract}
\section{Introduction}
\label{sec:intro}
It is often claimed that the performance of deep learning algorithms have attained super-human capabilities.
% This might be true, in a narrow extent, for data science domain where deep learning has led to the development of applications like Netflix's movie recommendation engine, or Facebook's news feed etc.
% The prediction mistakes made in such applications either do not have such catastrophic effects or they have not been studied yet e.g., the societal impacts of learned biases in Facebook's news feed.
However, real-world effectiveness of deep learning computer vision algorithms often fail to match the published performance on benchmarks \cite{Niko2018a}.
The vision models are largely fragile and do not generalize across realistic unconstrained scenarios \cite{Yin2019a}, e.g., when they encounter instances of classes, textures, or environmental conditions that were not covered by the training data \cite{Bijelic2020a}.
This could have serious consequences where the failures could lead to potentially catastrophic results as in the case of self-driving vehicles.

The authors of \cite{Hendrycks2020a} studied the existing hypotheses concerning the methods to improve the robustness of the deep networks.
Of particular relevance to our work, the technique of perturbing data without altering class labels, also known as data augmentation, has been proven to greatly improve model robustness and generalization performance \cite{Zoph2020a, Lopes2019a}.
It artificially inflates a dataset through label-preserving transforms to derive new samples from the originals.
Data augmentation offers straightforward strategies to learn invariances that are challenging to encode architecturally.
The most prevalent forms of data augmentation methodologies rely mostly on hand crafted features or include geometric distortions such as random cropping, zooming, rotation and flipping \cite{He2016a}.
Other methods include color space transformation, linear intensity scaling and elastic deformation.
These methods are successful at teaching mainly orientation and scale invariance, but are insufficient for other practical cases like occlusion, texture and complex illumination variations.
This is however often achieved by domain randomization on rendered images in simulation for Sim2Real transfer \cite{Tobin2017a}.
\begin{figure}[t]
	\centering
	\includegraphics[width=0.77\linewidth, trim={0cm 0cm 0cm 0cm},clip]{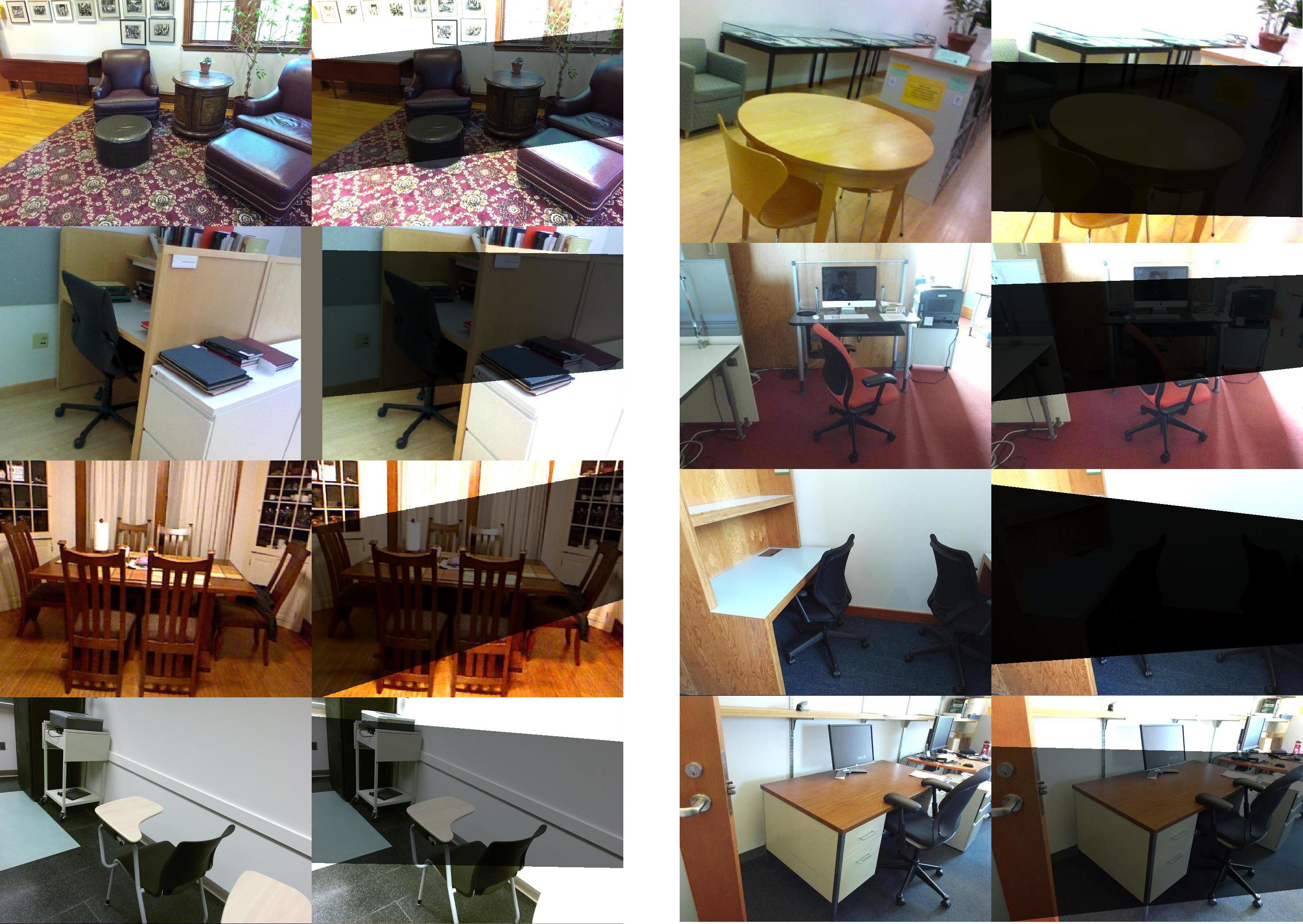}
	\caption{Samples of our proposed Random Shadows and Highlights (RSH) data augmentation scheme.
}
	\label{fig:RSH_Samples}
\end{figure}

Extreme lighting conditions such as dark shadows, intense highlights or change in brightness substantially affect the performance of deep models in image classification or object detection tasks.
To address such illumination variations and bridging the reality gap, this paper introduces a new data augmentation methodology, Random Shadows and Highlights (RSH) for real camera images.
The proposed strategy mimics high-contrast extreme lighting conditions by creating random shadows and highlights in the images.
% It can be applied both at image level or at patch level.
RSH is a parameter-learning free lightweight method which can be integrated with most vision related learning models without the need to change the learning strategy. The proposed data augmentation method is utilized to imitate the limitations of conventional RGB sensors to handle extreme light conditions.

\section{Related Works}
\label{sec:related_works}
Data augmentation is a strategy that aims to improve robustness of the models to unforeseen data shifts which they might encounter in real-world.
It explicitly teaches invariance to whichever transformation is used on the dataset.
For example, robustness against occlusions can be learned through Cutout \cite{DeVries2017a} and Random Erasing \cite{Zhong2020a}. 
Cutout randomly masks out square regions of input to generate new images to simulate occluded examples.
% This also guides the model to learn to take more of the image context into consideration when making decisions.
Random Erasing additionally varies size and aspect ratio of the masked region.
It also allows to replace pixel values with random noise.
Instead of occluding an image, CutMix \cite{Yun2019a} substitutes a portion of image with a portion of another image while the ground truth labels are also mixed proportionally to the area of the patches.
% This improves model robustness against input corruptions as well as its out-of-distribution performance.
Mixup \cite{Zhang2017a} is a another regularization strategy that produces elementwise convex combination of two images.
AugMix \cite{Hendrycks2019a} mixes together the results of several shorter augmentation chains in convex combinations to prevent image degradation while maintaining augmentation diversity.
% It couples Jensen-Shannon divergence consistency loss to enforce a smoother neural network response.
Alpha-blending two images generate pixel-level features that a camera might never produce and that could potentially affect model performance.
To prevent this, RICAP \cite{Takahashi2019a} spatially blend four cropped images to create a new image.
It performs occupancy estimation instead of classification, by mixing the four class labels with ratios proportional to the areas of the cropped images.
The method proposed in \cite{Sakkos2019a} is closest to our proposed strategy.
Nevertheless, they only create illumination circles on the images which does not represent the real world lighting perturbations often seen in indoor environments or even outdoors caused by shadows of the buildings.
Separate from hand-crafted methodologies are learned augmentation methods such as AutoAugment \cite{Cubuk2018a}, patch Gaussian augmentation \cite{Lopes2019a}, learned image transformations \cite{Zoph2020a} and DeepAugment \cite{Hendrycks2020a}.

\section{DATASETS}
\label{sec:datasets}
For \textbf{image classification}, we evaluate RSH on Tiny-ImageNet \cite{Le2015a}, CIFAR-10 and CIFAR-100 \cite{Krizhevsky2009a}.
Tiny-ImageNet is a miniature version of the ImageNet dataset with 200 classes.
Each class contains 500 training and 50 validation images yielding 100,000 training and 10,000 validation samples.
All images are of size $64\times64\times3$.
CIFAR-10 consists of 60,000 colour images in 10 classes with images of objects like airplanes, birds, trucks, etc.
Each class consists of 6,000 images.
The training set contains 50,000 images while 10,000 images are in the validation set.
CIFAR-100 is similar to CIFAR-10 except that it has 100 classes with 600 images in each class.
Each image comes with a fine and a coarse label.
% Fine labels are more descriptive than the coarse labels which actually refer to a super-class each label belongs to.
% For \textbf{object classification}, we utilize the FLIR Thermal Dataset \cite{Flir} which provides 8,862 train and 1,366 test images recorded in the streets and highways in Santa Barbara, California, USA.
% % The dataset provides unannotated, non-aligned RGB images as well for each thermal frame.
% The thermal images in the dataset are annotated with mainly four classes namely People, Bicycle, Car, Dog.
% The annotations are provided in the MSCOCO format.
For \textbf{object detection}, we utilize the PASCAL VOC 2007 dataset \cite{Everingham2007a} which provides 9,963 images of realistic scenes containing 24,640 annotated objects.
The dataset is split into 50\% for training/validation and 50\% for testing.
There are 20 classes of objects which are almost equally distributed in the images.

\section{OUR APPROACH}
\label{sec:our_approach}
Shadows cast by buildings in outdoors often take the form of trapezoids.
Furthermore, luminous light entering through windows or doors inside the buildings also take similar shapes.
We devise an algorithm to create Random Shadows and Highlights on images imitating trapezoidal shape with its base aligned with the vertical axis of the image.
\begin{algorithm}
    % \SetKwInOut{Input}{Input}
    \SetKwInOut{Output}{Output}
    \SetKwInput{kwInit}{Initialization}
    % \textbf{function} drawFilledPolygon $(p_1, p_2, p_3, p_4)$\;
    \KwIn{Input image ${I}$;}
    \nonl \myinput{Image size ${W}$, ${H}$ and $c$;}
    \nonl \myinput{RSH probability ${p}$;}
    \nonl \myinput{Highlights range ${H_{l}}$,  ${H_{h}}$;}
    \nonl \myinput{Shadows range ${S_{l}}$,  ${S_{h}}$;}
    \nonl \myinput{Left edge ranges ${l_{ul}}$, ${l_{uh}}$ and ${l_{ll}}$, ${l_{lh}}$;}
    \nonl \myinput{Right edge ranges ${r_{ul}}$, ${r_{uh}}$ and ${r_{ll}}$, ${r_{lh}}$.}
    \KwOut{Transformed image $I^*$}
    \kwInit{$p_1 \leftarrow \text{Rand}(0, 1).$}
    \eIf{$p_1 \geq p$}
      {
        $I^* \leftarrow I$; \\
        \Return $I^*$.
      }
      {
        $L_{lf} \leftarrow \text{Rand}(l_{ll}, l_{lh}) \times H;$ \\
        $L_{hf} \leftarrow \text{Rand}(l_{ul}, l_{uh}) \times H;$ \\
        $R_{lf} \leftarrow \text{Rand}(r_{ll}, r_{lh}) \times H;$ \\
        $R_{hf} \leftarrow \text{Rand}(r_{ul}, r_{uh}) \times H;$ \\
        $P_{bl} \leftarrow (0, L_{lf} + L_{hf});$ \\
        $P_{tl} \leftarrow (0, L_{hf});$ \\
        $P_{br} \leftarrow (W, R_{lf} + R_{hf});$ \\
        $P_{tr} \leftarrow (W, R_{hf});$ \\
        $I_{mask} \leftarrow \text{ContourFill}(\text{Zeros}(H, W, c), [P_{tl}, P_{tr}, P_{br}, P_{bl}])$; \\
        $I_{inv\_mask} \leftarrow \neg I_{mask}$; \\
        $S_f \leftarrow \text{Rand}(S_l, S_h)$; \\
        $H_f \leftarrow \text{Rand}(H_l, H_h)$; \\
        $I_{shadow} \leftarrow \text{adjustBrightness}(I, S_f)$; \\
        $I_{high} \leftarrow \text{adjustBrightness}(I, H_f)$; \\
        $I^* \leftarrow I_{shadow}(I_{mask}) + I_{high}(I_{inv\_mask})$; \\
        \Return $I^*.$

      }
    \caption{Random Shadows and Highlights}
    \label{alg:RSH}
\end{algorithm}
For an image ${I}$, Random Shadows and Highlights is applied with a probability $p$ in training.
For probability $1-p$, the image is left unchanged.
RSH is driven by twelve input parameters to create random shadows and highlights.
These variables include highlights range $H_{l}$, $H_{h}$ and shadows range $S_{l}$, $S_{h}$.
%to determine pixel intensities of highlights and shadows in the output image.
These also include left edge ranges $l_{ul}$, $l_{uh}$ and $l_{ll}$, $l_{lh}$, and right edge ranges $r_{ul}$, $r_{uh}$ and $r_{ll}$, $r_{lh}$, to randomly obtain four points in total i.e., two on each vertical axis of the image which are subsequently employed to draw a trapezoid.
For an input image of size $W \times H$, our algorithm
% When an image of size $W \times H$ is passed to RSH, it
randomly
creates a contour on the image driven by 
the provided parameters ranges.
The brightness of the pixels inside the contour is reduced by a factor $S_f$ which is randomly chosen within the predefined range $S_l$ and $S_h$.
Similarly, the brightness of the pixels outside the contour is increased by a factor $H_f$ obtained from within $H_l$ and $H_h$.
The detailed procedure of creating the contour mask(s) and ultimately creating an image with shadows and highlights is shown in Alg~\ref{alg:RSH}.

\section{EXPERIMENTS}
\label{sec:experiment}
\subsection{Image Classification}
\label{ssec:image_classification}
% \subsubsection{Experiment Settings}
% \label{sssec:experiment_setting}

Through extensive experimentation, we study the impact of Random Shadows and Highlights (RSH) in comparison to a set of related augmentations methods.
% Existing similar strategies in the literature often utilize a combination of other regularization and data augmentation methods to improve their scores in benchmarks.
% Conversely, we do not employ any such additional method and focus precisely on studying the improvements brought by our proposed strategy.
Among the commonly employed strategies, the most relevant ones to our work are \textit{Gamma Correction} and \textit{Color Jitter}.
Varying powers of gamma darkens or lightens the shadows in the image.
While in Color Jitter, we randomly change the brightness, contrast, saturation and hue of the image.
Moreover, we also compare our results with the method proposed in \cite{Sakkos2019a}, we call it \textit{Disk Illumination}.
To study over-fitting, we quantify the Train-Test Difference (TTD), as the name suggests, difference in error on the train set and test set, with and without RSH perturbations.
When lighting corruption is applied to the test set, its probability is set to $1$.
The RSH parameters which were chosen in our experiments are: $H_{l}=1$, $H_{h}=2$, $S_{l}=0$, $S_{h}=1$, $l_{ul}$ and $r_{ul}=0$, $l_{uh}$ and $r_{uh}=0.3$, $l_{ll}$ and $r_{ll}=0.4$, $l_{lh}$ and $r_{lh}=0.8$.
Gamma is randomly changed from $0$ to $1.5$ while brightness, contrast and saturation ranges are $0$ to $2$, and hue changes from $-0.5$ to $0.5$ randomly.
% In our experiments, the performance of different CNN architectures with or without Random Shadows and Highlights is evaluated.
% However, we do compare our approach against a set of similar existing data augmentation methods in our experiments.

\textbf{Architectures: } We adopted two architectures on Tiny-ImageNet, CIFAR-10 and CIFAR 100: EfficientNet \cite{Tan2019a} and AlexNet \cite{Krizhevsky2017a}.
More precisely, we employed the EfficientNet-B0 architecture.
The models are pre-trained on the ImageNet dataset except the \textit{SoftMax} layer which is adopted for each dataset and initialized with random weights.
The training is performed only for $20$ epochs at a learning rate of $1e^{-3}$ and a momentum of $0.9$ with the \textit{SGD} optimizer.
% When training for CIFAR-10 and CIFAR-100, images are resized to the size $64 \times 64 \times 3$.
\begin{table}[t]
\centering
\begin{tabular}{c|c|c|c|c|c} 
\hline
Method       & \begin{tabular}[c]{@{}c@{}}Train\\Error~\end{tabular} & \begin{tabular}[c]{@{}c@{}}Test\\Error~\end{tabular} & \begin{tabular}[c]{@{}c@{}} Test\\Error\\RSH (1) \end{tabular} & \begin{tabular}[c]{@{}c@{}}TTD\\w/o\\RSH\end{tabular}   & \begin{tabular}[c]{@{}c@{}}TTD\\RSH\\(1)\end{tabular}  \\ 
\hline\hline
Baseline           & 0.303                                                 & 0.408                                                & 0.630                                                      & 0.105  & 0.327                                                    \\ 
\hdashline
RGC (0.5)          & 0.301                                                 & 0.396                                                & 0.603                                                      & 0.095      & 0.302                                                        \\
RCJ (0.5)         & 0.415                                               & 0.406                                                & 0.612                                                      & $-$0.009 & 0.197                                                    \\
RDI (0.5)          & 0.493                                                 & 0.410                                                & 0.627                                                      & $-$0.083  & 0.134                                                    \\
\textbf{RSH} (0.5) & 0.353                                                 & 0.407                                                & \textbf{0.449}                                             & \textbf{0.054}  & \textbf{0.096}                                                    \\ 
\hdashline
RGC (1)         & 0.323                                                 & 0.400                                                & 0.600                                                      & 0.077      & 0.277                                                        \\
RCJ (1)         & 0.485                                                 & 0.459                                                & 0.656                                                      & $-$0.026 & 0.171                                                    \\
RDI (1)          & 0.639                                                 & 0.631                                              & 0.780                                                      & $-$0.008  & 0.141                                                    \\
\textbf{RSH} (1) & 0.411                                                 & 0.434                                                & \textbf{0.450}                                             & \textbf{0.023}  & \textbf{0.039}                                                    \\
\hline
\end{tabular}
\caption{\label{tab:tiny-imagenet-effnetb0-error} Performance evaluation of EfficientNet-B0 on TinyImageNet. For comparison, models are trained with Random Gamma Correction (RGC), Random Color Jitter (RCJ) and Random Disk Illumination (RDI) as well. Apart from train and test error computation, performance is measured with lighting perturbations as well by applying RSH in the test set with $p=1$. Moreover, the Train-Test Difference (TTD) is also computed to quantify over-fitting of the models. Negative TTD implies over-regularization which is not desired.
}
\vspace{-0.5cm}
\end{table}
\subsubsection{Classification Evaluation}
\label{sssec:class_eval}
\begin{figure}[h]
	\centering
	\includegraphics[width=0.85\linewidth, trim={0.3cm 0.4cm 0.4cm 0.4cm},clip]{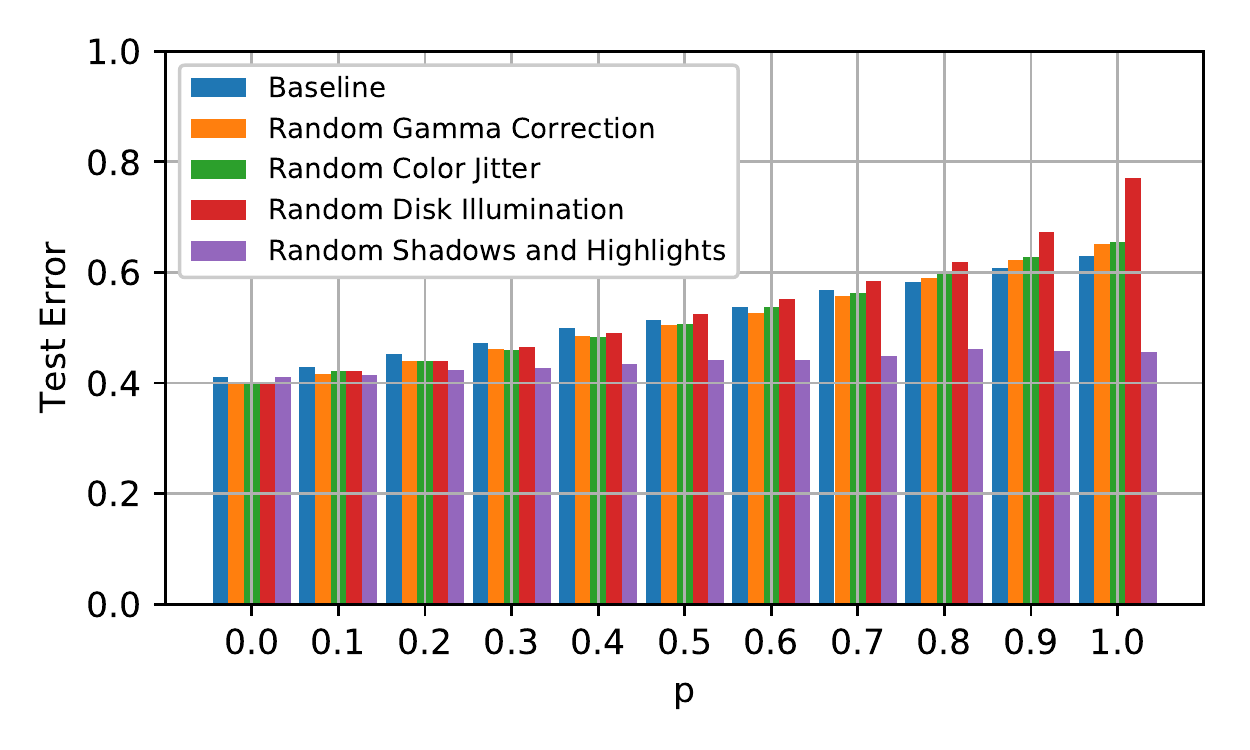}
	\caption{Test error comparison between different data augmentation strategies for different levels of lighting perturbations. \textit{p} represents data augmentation activation probabilities for each individual technique in training as well as that of lighting perturbations in test set.}
	\label{fig:Test_Error_all}
\end{figure}

The performances of the selected data augmentation strategies are evaluated at different activation probabilities $p$, ranging from $0.0$ to $1.0$ with a step size of $0.1$.
Table~\ref{tab:tiny-imagenet-effnetb0-error} presents the results of these experiments at $p=0.5$ and $p=1$ only. 
At a first glance, it may appear that the performance with RSH decreases slightly on test sets when no lighting perturbations are applied.
To some extent, this is true and is justified by the observations made in \cite{Tsipras2018a} that argues that standard accuracy is often compromised to obtain robust models.
However, when tested with lighting corruptions, our strategy out-scores all other methods with a considerable margin by obtaining least test errors at both probabilities.
In fact, Figure~\ref{fig:Test_Error_all} illustrates that RSH is extremely robust against lighting perturbations at all values of $p$.
Table~\ref{tab:tiny-imagenet-effnetb0-error}, also shows that for lighting perturbations at $p=1$ on the test set, the performance of the model trained with RSH is increased by approximately 25\% at both $p=0.5$ and $p=1$ when compared to RGC which performed best among other strategies.
Furthermore, we estimate the Train-Test Difference (TTD) to measure over-fitting for cases with and without lighting corruption.
In addition to the increased robustness, 
our RSH augmentation method proves to have reduced over-fitting significantly as well.

\begin{figure}[t]
	\centering
	\includegraphics[width=0.8\linewidth, trim={0cm 0cm 0cm 0cm},clip]{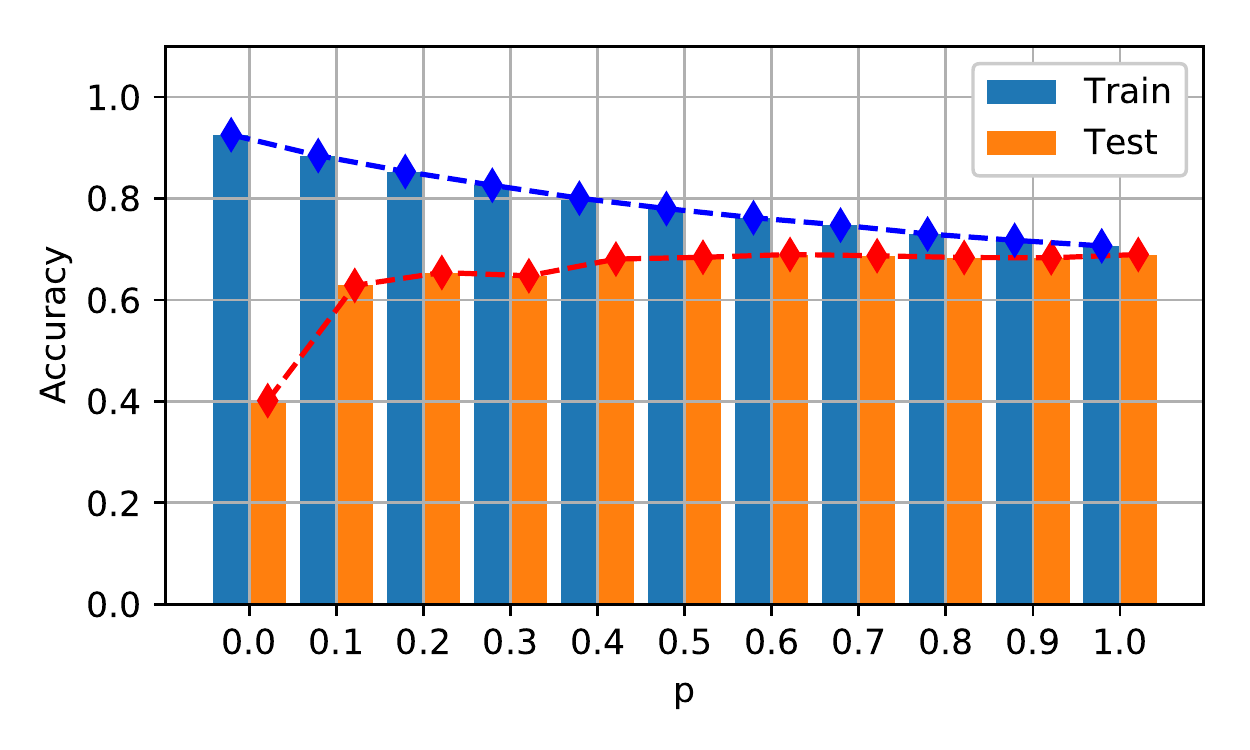}
% 	\vspace{-0.4cm}
	\caption{Performance evaluation on CIFAR-10 when trained with different activation probabilities of RSH. The probability of lighting corruptions is set to $1$ for all values of $p$. It is evident that with the increase in $p$, over-fitting decreases continuously until the model fits almost perfectly for both train and test sets.}
	\label{fig:hyper-p}
\end{figure}

\textbf{Impact of hyperparameters:}
We also conduct experiments to study the impact of RSH hyperparameters on the model performance.
Figure~\ref{fig:hyper-p} illustrates the results obtained when training AlexNet on CIFAR-10 with increasing RSH probability from $0.0$ to $1.0$.
The probability of lighting corruption is set to 1 in the test set.
It is evident, that the model overfits severely when no RSH augmentation is applied during training.
Nevertheless, RSH $p$ increases in the training, robustness of the network increases continuously until we obtain almost the same train-test prediction accuracy, which is desired in most cases.
\begin{figure}[b]
	\centering
	\includegraphics[width=0.85\linewidth, trim={0cm 0cm 0cm 0cm},clip]{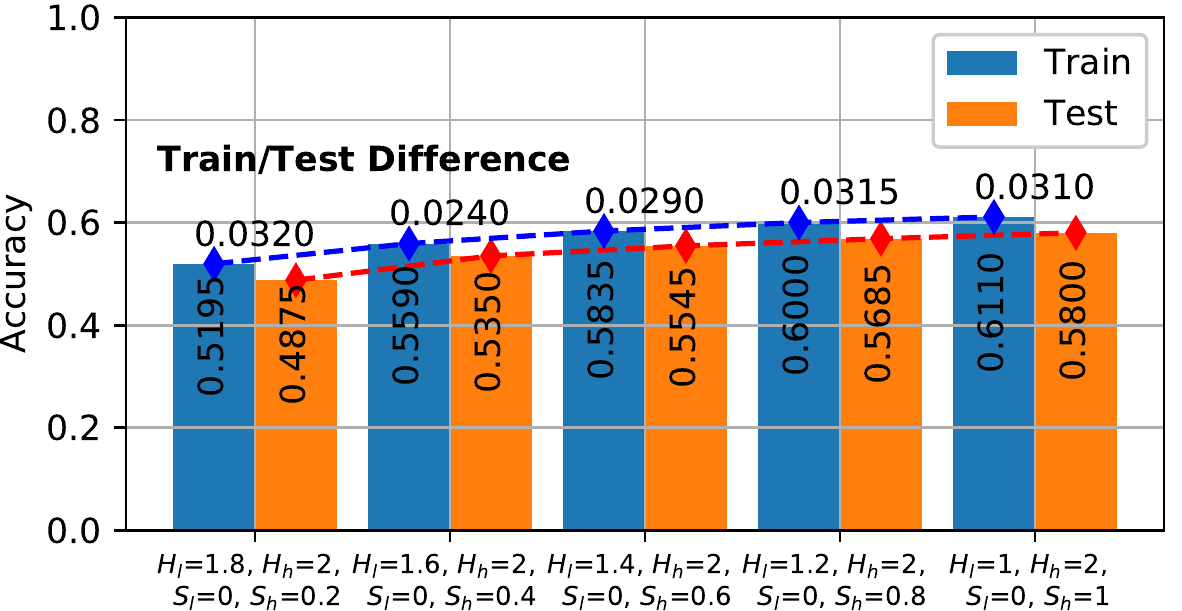}
% 	\vspace{-0.4cm}
	\caption{Impact of changing $H$ and $S$ hyperparameters. For all cases we set $H_h=2$ while $S_l=0$. The X axis represents increasing span of choice for highlights and shadows. Clearly, with wider range, RSH performs best in terms of accuracy.
	%TTD is also shown which suggest slight decrease and then increase as we move along horizontal axis. When compared to the choice with $H_l=1.8$ and $S_h=0.2$ (i.e., the most narrow $H$ and $S$ range), TTD of the best performing (last) set of parameters has improved by 3.125\%.
	}
	\label{fig:impact_of_h_s}
\end{figure}
We also evaluate the impact of changing other hyperparameters.
Figure~\ref{fig:impact_of_h_s} illustrates that increasing the range of highlights and shadows improves not only the accuracy of the model, but enhances the TTD by 3.125\% as well.
The significance of the ``shadows area'' is also assessed on CIFAR-100 by changing the values of $l_{lh}, r_{lh}, l_{ul}$ and $r_{ul}$ while keeping $l_{ll}, r_{ll}, l_{uh}$ and $r_{uh}$ constant.
A marginal improvement of 1.7\% is observed with the values presented in Section~\ref{ssec:image_classification} as compared to the the least performing ones.

\subsection{Object Detection}
\label{ssec:Object Detection}
In our experiments for object detection, we employ a single-stage end-to-end object detector DETR proposed in \cite{Carion2020a}.
It is among the pioneering works that exploited transformers in the image domain.
However, DETR technically is still a combination of CNN and self-attention.
We utilize ResNet-50 as the backbone features extractor while the object detector head is adopted to suit the number of classes in the Pascal VOC 2007 dataset.
The weights are initialized from a checkpoint pretrained on the COCO object dataset.
Thus, the hyperparameters of the transformer in DETR are set to default with $6$ encoder and decoder layers, $8$ attention heads, and $100$ number of queries.
The learning rate for both backbone and transformer layers is set to $8 \times 10^{-6}$ while training is terminated after $150$ epochs for each experiment.
% As described in Section~\ref{sec:datasets}, FLIR-Thermal dataset only provides annotated thermal images while their RGB counterparts are neither aligned nor annotated.
% We exploit the concept of homography and manually select the matching features in a set of RGB and thermal images.
% A transformation matrix is obtained which we use to align RGB images with the corresponding thermal images.
We perform experiments with and without Random Shadows and Highlights augmentation on the Pascal VOC 2007 dataset.
\begin{table}[t]
\centering
\begin{tabular}{c|c|c|c|c|c} 
\hline
\multirow{2}{*}{Model} & \multicolumn{3}{c|}{~mAP@IoU=0.5}                                             & \multirow{2}{*}{\begin{tabular}[c]{@{}c@{}}TTD\\w/o\\RSH\end{tabular}} & \multirow{2}{*}{\begin{tabular}[c]{@{}c@{}}TTD w/\\RSH\\(1)\end{tabular}}  \\ 
\cline{2-4}
                       & Train & Test           & \begin{tabular}[c]{@{}c@{}}Test w/\\RSH (1)\end{tabular} &                                                                       &                                                                       \\ 
\hline\hline
Baseline               & 0.888 & 0.751          & 0.639                                                & 0.137                                                                 & 0.249                                                                 \\ 
\hline
\textbf{RSH (1)}           & 0.872 & \textbf{0.751} & \textbf{0.705}                                       & \textbf{0.121}                                                        & \textbf{0.167}                                                        \\
\hline
\end{tabular}
\caption{\label{tab:FLIR}Performance evaluation of our RSH augmentation strategy for object detection on the Pascal VOC 2007 dataset.}
\end{table}

% \vspace{-0.4cm}
\subsubsection{Detection Evaluation}
\label{sssec:detection_eval}
The performance evaluation of DETR with RSH on the Pascal VOC 2007 dataset is presented in Table~\ref{tab:FLIR}.
For each setting, mean of 20 evaluation trials is computed.
It is evident from the results that the model trained with our RSH augmentation not only maintained its standard accuracy but also improved the robustness against lighting corruption by 10.328\% compared to the baseline.
% In this case, not only the standard accuracy of the object detector has increased by 1.417\% with our data augmentation method, but also the robustness of the detector against lighting corruptions by 5.017\% compared to the baseline respectively.
Clear improvements in TTD with and without RSH perturbations in the test set can also be seen.

\section{CONCLUSION}
\label{sec:conclusion}
A new lightweight learning-free data augmentation strategy is presented in this paper to acquire robustness against lighting corruption.
This is achieved by creating random shadows and highlights on the images during training.
The models trained with the proposed augmentation method not only out-scored other similar methods against lighting perturbations but also reduced over-fitting on the training data significantly.
The results in the performed experiments demonstrate that RSH should be considered essential in CNN model training.
In the future, we plan to apply our method on other applications, e.g, facial recognition and person re-identification.

%\vfill
%\pagebreak

\bibliographystyle{IEEEbib}
\bibliography{main}

\begin{thebibliography}{10}

\bibitem{Niko2018a}
N.~S{\"u}nderhauf, O.~Brock, W.~Scheirer, R.~Hadsell, D.~Fox, J.~Leitner,
  B.~Upcroft, P.~Abbeel, W.~Burgard, M.~Milford, et~al.,
\newblock ``The limits and potentials of deep learning for robotics,''
\newblock {\em The Intl. Journal of Robotics Research}, vol. 37, no. 4-5, pp.
  405--420, 2018.

\bibitem{Yin2019a}
D.~Yin, R.~Gontijo Lopes, Jon S., Ekin~D. C., and J.~Gilmer,
\newblock ``A {F}ourier perspective on model robustness in computer vision,''
\newblock in {\em Advances in Neural Information Processing Systems}, 2019, pp.
  13276--13286.

\bibitem{Bijelic2020a}
M.~Bijelic, T.~Gruber, F.~Mannan, F.~Kraus, W.~Ritter, K.~Dietmayer, and
  F.~Heide,
\newblock ``Seeing through fog without seeing fog: Deep multimodal sensor
  fusion in unseen adverse weather,''
\newblock in {\em Proc. of the IEEE/CVF Conf. on Computer Vision and Pattern
  Recognition}, 2020, pp. 11682--11692.

\bibitem{Hendrycks2020a}
D.~Hendrycks, S.~Basart, N.~Mu, S.~Kadavath, F.~Wang, E.~Dorundo, R.~Desai,
  T.~Zhu, S.~Parajuli, M.~Guo, et~al.,
\newblock ``The many faces of robustness: A critical analysis of
  out-of-distribution generalization,''
\newblock {\em arXiv preprint arXiv:2006.16241}, 2020.

\bibitem{Zoph2020a}
B.~Zoph, E.~D. Cubuk, G.~Ghiasi, T.~Lin, J.~Shlens, and Q.~V. Le,
\newblock ``Learning data augmentation strategies for object detection,''
\newblock in {\em European Conf. on Computer Vision}. Springer, 2020, pp.
  566--583.

\bibitem{Lopes2019a}
R.~G. Lopes, D.~Yin, B.~Poole, J.~Gilmer, and E.~D. Cubuk,
\newblock ``Improving robustness without sacrificing accuracy with patch
  {G}aussian augmentation,''
\newblock {\em arXiv preprint arXiv:1906.02611}, 2019.

\bibitem{He2016a}
K.~He, X.~Zhang, S.~Ren, and J.~Sun,
\newblock ``Deep residual learning for image recognition,''
\newblock in {\em Proc. of the IEEE Conf. on Computer Vision and Pattern
  Recognition}, 2016, pp. 770--778.

\bibitem{Tobin2017a}
J.~Tobin, R.~Fong, A.~Ray, J.~Schneider, W.~Zaremba, and P.~Abbeel,
\newblock ``Domain randomization for transferring deep neural networks from
  simulation to the real world,''
\newblock in {\em 2017 IEEE/RSJ Intl. Conf. on Intelligent Robots and Systems
  (IROS)}. IEEE, 2017, pp. 23--30.

\bibitem{DeVries2017a}
T.~DeVries and G.~W Taylor,
\newblock ``Improved regularization of convolutional neural networks with
  cutout,''
\newblock {\em arXiv preprint arXiv:1708.04552}, 2017.

\bibitem{Zhong2020a}
Z.~Zhong, L.~Zheng, G.~Kang, S.~Li, and Y.~Yang,
\newblock ``Random erasing data augmentation,''
\newblock {\em Proc. of the AAAI Conf. on Artificial Intelligence}, vol. 34,
  no. 07, pp. 13001--13008, Apr. 2020.

\bibitem{Yun2019a}
S.~Yun, D.~Han, S.~J. Oh, S.~Chun, J.~Choe, and Y.~Yoo,
\newblock ``{CutMix}: Regularization strategy to train strong classifiers with
  localizable features,''
\newblock in {\em Proc. of the IEEE Intl. Conf. on Computer Vision}, 2019, pp.
  6023--6032.

\bibitem{Zhang2017a}
H.~Zhang, M.~Cisse, Y.~N. Dauphin, and D.~Lopez-Paz,
\newblock ``mixup: Beyond empirical risk minimization,''
\newblock {\em arXiv preprint arXiv:1710.09412}, 2017.

\bibitem{Hendrycks2019a}
D.~Hendrycks, N.~Mu, E.~D. Cubuk, B.~Zoph, J.~Gilmer, and B.~Lakshminarayanan,
\newblock ``Augmix: A simple data processing method to improve robustness and
  uncertainty,''
\newblock {\em arXiv preprint arXiv:1912.02781}, 2019.

\bibitem{Takahashi2019a}
R.~Takahashi, T.~Matsubara, and K.~Uehara,
\newblock ``Data augmentation using random image cropping and patching for deep
  {CNN}s,''
\newblock {\em IEEE Trans. on Circuits and Systems for Video Technology}, 2019.

\bibitem{Sakkos2019a}
D.~Sakkos, H.~P. Shum, and E.~S. Ho,
\newblock ``Illumination-based data augmentation for robust background
  subtraction,''
\newblock in {\em 13th Intl. Conf. on Software, Knowledge, Information
  Management and Applications (SKIMA)}. IEEE, 2019, pp. 1--8.

\bibitem{Cubuk2018a}
E.~D. Cubuk, B.~Zoph, D.~Mane, V.~Vasudevan, and Q.~V. Le,
\newblock ``{AutoAugment}: Learning augmentation policies from data,''
\newblock {\em arXiv preprint arXiv:1805.09501}, 2018.

\bibitem{Le2015a}
Y.~Le and X.~Yang,
\newblock ``{Tiny ImageNet Visual Recognition Challenge},''
  \url{https://tiny-imagenet.herokuapp.com/}, 2015.

\bibitem{Krizhevsky2009a}
A.~Krizhevsky et~al.,
\newblock ``Learning multiple layers of features from tiny images,''
\newblock {\em Tech Report}, 2009.

\bibitem{Everingham2007a}
M.~Everingham, L.~Van~Gool, C.~K.~I. Williams, J.~Winn, and A.~Zisserman,
\newblock ``The {PASCAL} {V}isual {O}bject {C}lasses {C}hallenge 2007
  {(VOC2007)} {R}esults,''
  http://www.pascal-network.org/challenges/VOC/voc2007/workshop/index.html.

\bibitem{Tan2019a}
M.~Tan and Q.~V. Le,
\newblock ``{EfficientNet}: Rethinking model scaling for convolutional neural
  networks,''
\newblock {\em arXiv preprint arXiv:1905.11946}, 2019.

\bibitem{Krizhevsky2017a}
A.~Krizhevsky, I.~Sutskever, and G.~E. Hinton,
\newblock ``{ImageNet} classification with deep convolutional neural
  networks,''
\newblock {\em Communications of the ACM}, vol. 60, no. 6, pp. 84--90, 2017.

\bibitem{Tsipras2018a}
D.~Tsipras, S.~Santurkar, L.~Engstrom, A.~Turner, and A.~Madry,
\newblock ``Robustness may be at odds with accuracy,''
\newblock {\em arXiv preprint arXiv:1805.12152}, 2018.

\bibitem{Carion2020a}
N.~Carion, F.~Massa, G.~Synnaeve, N.~Usunier, A.~Kirillov, and S.~Zagoruyko,
\newblock ``End-to-end object detection with transformers,''
\newblock {\em arXiv preprint arXiv:2005.12872}, 2020.

\end{thebibliography}

\end{document}